\documentclass[runningheads]{llncs}
\usepackage{graphicx}
\usepackage{stfloats}
\usepackage{amsmath}
\usepackage{amsfonts}
\usepackage{cleveref}
\usepackage{booktabs}
\usepackage{wrapfig}
\usepackage{lipsum}
\usepackage{authblk}

\begin{document}
\title{Generation of Differentially Private Heterogeneous Electronic Health Records}
\titlerunning{Generation of DP Heterogeneous EHRs}
%
\author{Kieran Chin-Cheong\inst{1} \and
Thomas Sutter\inst{1}\ \and
Julia E. Vogt\inst{1}}
\authorrunning{K. Chin-Cheong et al.}
%
\institute{Department of Computer Science, ETH Zurich, Zurich, Switzerland\\
\email{\{kieran.chincheong,thomas.sutter,julia.vogt\}@inf.ethz.ch}}

\maketitle              

\begin{abstract}
  Electronic Health Records (EHRs) are commonly used by the machine learning community for research on problems specifically related to health care and medicine. EHRs have the advantages that they can be easily distributed and contain many features useful for e.g. classification problems. What makes EHR data sets different from typical machine learning data sets is that they are often very sparse, due to their high dimensionality, and often contain heterogeneous (mixed) data types. Furthermore, the data sets deal with sensitive information, which limits the distribution of any models learned using them, due to privacy concerns. For these reasons, using EHR data in practice presents a real challenge. In this work, we explore using Generative Adversarial Networks to generate synthetic, \textit{heterogeneous} EHRs with the goal of using these synthetic records in place of existing data sets for downstream classification tasks. We will further explore applying differential privacy (DP) preserving optimization in order to produce DP synthetic EHR data sets, which provide rigorous privacy guarantees, and are therefore shareable and usable in the real world. The performance (measured by AUROC, AUPRC and accuracy) of our model's synthetic, heterogeneous data is very close to the original data set (within 3 - 5\% of the baseline) for the non-DP model when tested in a binary classification task. Using strong $(1, 10^{-5})$ DP, our model still produces data useful for machine learning tasks, albeit incurring a roughly 17\% performance penalty in our tested classification task. We additionally perform a sub-population analysis and find that our model does not introduce any bias into the synthetic EHR data compared to the baseline in either male/female populations, or the 0-18, 19-50 and 51+ age groups in terms of classification performance for either the non-DP or DP variant.
\end{abstract}
\section{Introduction}
\label{sec:introduction}

Data sets used by the machine learning community for research on problems related to health care and medicine are often based on Electronic Health Records (EHRs). These records contain private details about patients' visits to hospitals or health-care facilities, and in particular usually consist of heterogeneous administrative data (such as patient age, weight, or length of stay), and diagnostic data (such as associated ICD codes \cite{icd10} for both diagnoses made and procedures carried out). The administrative data is typically dense, while the diagnostic data is typically very sparse. Together, this results in a data set that is heterogeneous, high dimensional, and sparse. Examples of such data sets are the freely accessible Mimic-III data set \cite{mimiciii} and the New Zealand National Minimal data set \cite{NZDataSet}.

A typical use case for medical data sets is to perform binary classification. One such task is to try to use EHRs to predict the risk of a patient being unexpectedly readmitted to hospital after being discharged. This is a well-studied and relevant clinical task, because unexpected hospital readmissions are financially penalized in several countries, by programs such as the Hospital Readmission Reduction Program in the United States \cite{hrrp}, which has cost hospitals \$1.9 billion in penalties as of 2016 \cite{aha}.

Using EHR data to train machine learning models requires some care, however, as the data being used is inherently sensitive. For instance, it is possible to recover training data from models \cite{Fredrikson:2015:MIA:2810103.2813677}. Furthermore, even though in many cases the training data has been de-identified, it has also been shown that data re-identification is possible via linkage to external data sources \cite{10.1371/journal.pone.0028071}. Therefore, there exists a risk to the privacy of patients whose data is present in training data sets. One way to avoid these pitfalls is to release models trained using synthetic data sets, which are based on the original training data. Such models might provide better privacy to patients, because training data extraction attacks would then have to further re-identify a synthetic training record. However, to truly provide rigorous privacy guarantees, the synthetic data should be generated in a differentially private (DP) \cite{Dwork:2014:AFD:2693052.2693053} manner.

In this work, we investigate the generation of heterogeneous EHR data for use in downstream classification tasks. In particular, we will use returning hospital patient classification as our evaluation task. However, this is certainly not the only applicable downstream task. We will further apply DP to the EHR generation and investigate the impact that this has, specifically on the training process itself and on the difference in quality between the DP and non-private generated EHR data. Finally, we will perform a sub-population analysis on the training data, and synthetic EHR data from both models in an attempt to determine whether or not our models introduce any bias in their synthetic data.

\section{Background and Related Work}
\label{sec:related_work}

\subsection{Generative Adversarial Networks}
\label{subsec:GANs}

In this work we will focus mainly on generative machine learning models. One such model which has recently enjoyed much success in many different domains is the Generative Adversarial Network (GAN), first proposed by Goodfellow et al \cite{goodfellow:2014}. GANs are a type of adversarial training system, where two competing models are trained against each other. The generator model attempts to transform random input noise into samples mirroring those of the training data distribution, and the discriminator model attempts to distinguish between real training data samples and generated samples. These two models are typically implemented as neural networks, and are optimized against each other, resulting in the following minimax game with value function $V(D, G)$:

\begin{equation}
    \min_{G}\max_{D} V(D, G) = \mathbb{E}_{x\sim p_{data(x)}}[\log D(x)]+ \mathbb{E}_{z\sim p_z(z)}[\log(1-D(G(z)))]
\end{equation}

where D represents the discriminator and G represents the generator. $D(x)$ outputs the probability that the sample $x$ has been drawn from the training data, and $G(z)$ is the generated sample given input noise $z$, where $z$ is typically sampled from either a uniform or Gaussian distribution. It has been theoretically shown that a global optimum exists where the generator network recovers the distribution of the training data set, that is: $p_{data} = p_{g}$ \cite{goodfellow:2014}. 

However, in practice it is difficult to train a GAN to optimality. In particular, balancing the power of the discriminator and generator networks is important, as otherwise the generator can learn to simply output similar samples regardless of the random noise input. This so-called "mode collapse" problem can be hard to avoid, and requires much ad-hoc parameter tuning. GANs can also suffer from vanishing gradients, particularly when the discriminator is very strong. In their paper, Salimans et al \cite{DBLP:journals/corr/SalimansGZCRC16} investigate and suggest several techniques to improve GAN training.

A different approach is to re-formulate the loss of the GAN in terms of the Wasserstein-1 distance between the generated and training data distributions \cite{arjovsky:2017}. The intuition here is that for many problems, using the Jenson-Shannon divergence to quantify the difference between generated and training data fails to provide a smooth and continuous gradient everywhere. However, using the Wasserstein-1 distance alleviates this problem. As a result, the so-called Wasserstein GANs (WGANs) do not seem to suffer from mode collapse \cite{arjovsky:2017}. Calculating the Wasserstein-1 distance is unfortunately intractable, so the authors propose to approximate it by instead learning a K-Lipschitz function maximizing the expected difference between the generated and training data distributions:

\begin{equation}
\max_{w \in W}\mathbb{E}_{x\sim \mathbb{P}_{data}}[f_{w}(x)]-\mathbb{E}_{z\sim\mathbb{P}_{p(z)}}[f_{w}(G(z))]
\end{equation}

where the functions $\left\lbrace f_{w}\right\rbrace_{w \in W}$ are all K-Lipschitz for some unknown value of K. The only remaining problem is how to enforce the K-Lipschitz constraint. The original WGAN paper clips the weights of the learned function when it is implemented as a neural network, and a follow-up paper instead uses a regularizer which penalizes the magnitude of the gradient of the loss function as it diverges from 1 \cite{DBLP:journals/corr/GulrajaniAADC17}. These GAN variants will be referred to as WGAN and WGAN-GP respectively in the rest of this work.

One recent, applicable line of research focuses on using GANs to generate tabular data. The CTGAN \cite{xu2019modeling}, is one such work. This model introduces two techniques to combat mode collapse when generating tabular data: mode-specific normalization and conditional training by sampling. While pixel intensities in images follow a Gaussian-like distribution, this is not always the case for other data types. In EHR data, features like age or length of stay often have multi-modal distributions, or are non-standard. Mode-specific normalization attempts to mitigate this problem by using a Gaussian Mixture Model to decompose such features into two new features: one indicating the mode within the original data that the sample belongs to, and a value indicating the value within that mode. Conditional training by sampling leverages a conditional GAN combined with a training sample selection regime, which aims to ensure that when randomly choosing training samples, classes with low representation are still selected.

\subsection{EHR Data Generation}

Several previous works have investigated the generation of EHR data using generative models like GANs. One such work is MedGAN, introduced in 2017 by Choi et al \cite{choi:2017}. MedGAN uses a combination of a GAN and an autoencoder to successfully generate EHR data. This paper also introduced several useful techniques which are necessary to help the GAN converge during training, which are particularly useful for EHR data sets. However, MedGAN is unable to handle data sets which contain heterogeneous data types, such as data sets including continuous, binary and count variables. Instead, datasets contain only either binary or count variables, but not both.

Baowaly et al \cite{10.1093/jamia/ocy142} apply the WGAN and WGAN-GP modifications previously discussed to the MedGAN framework, similar to we will do in this work. However, like MedGAN, their model only focuses on diagnosis codes or counts, and does not include heterogeneous data types or administrative data.

Esteban et al \cite{esteban2017realvalued} investigate using recurrent and recurrent conditional GANs to generate real-valued medical time series data also used for a downstream task, in this case training an early warning system based on four synthetically generated real-valued features. Similarly, Beaulieu-Jones et al \cite{doi:10.1161/CIRCOUTCOMES.118.005122} use an Auxiliary Classifier GAN (AC-GAN) to generate time series data consisting of 3 repeatedly measured parameters based on the SPRINT clinical trial. Both of these works focus on the generation of a small number of real-valued features, and do not include administrative or diagnosis data.

Our contribution in this work differs from such previous works in that our model is capable of generating heterogeneous EHR data, i.e. binary, count and real-valued data at the same time.

\subsection{Patient Readmission Classification}
\label{subsec:readmission_classification}

Machine learning and statistical methods have also been applied to the task of hospital patient readmission classification, an important clinical problem as patients become re-hospitalized shortly after their release. In addition, there are costs associated with early readmissions \cite{aha}. Several various models and techniques have already been employed. He et al \cite{f716917a75c346a8b6dd0c5f22cac0b0} use multivariate logistic regression to classify patients at risk of readmission over their dataset and achieve an AUROC value of up to 0.81 in the best case. Futoma et al \cite{FUTOMA2015229} evaluate several classifiers, including deep feed-forward networks, and obtain varying AUROC values up to 0.734 depending on the patient disease category.

Xiao et al \cite{10.1371/journal.pone.0195024} take a slightly different approach, where they learn an interpretable patient representation using EHR data using a hybrid Topic Recurrent Neural Network, and then use this to predict patient readmission for patients with Congestive Heart Failure. Here, they obtain an AUROC of 0.61, an AUPRC of 0.39 and an accuracy of 0.69.

The baseline classifiers in our work match or exceed the classification performance obtained in these works, and classifiers trained using our non-private generated data performs at a similar level.

\subsection{Differential Privacy}
\label{subsec:differential_privacy}

Machine learning is becoming more prevalent, and at the same time more and more sensitive data is collected and being used as training data for models. This naturally creates a need to ensure that this data is used responsibly, and respects the privacy of the individuals in question. One way to try to achieve these goals is to apply DP \cite{10.1007/11681878_14} either when training machine learning models, or when releasing models to the public based on sensitive data. Intuitively, DP ensures that the presence or absence of a particular record in a data set has a very small impact on the probability of a particular outcome of an algorithm. This therefore protects privacy, as any information gained about an individual must be independent of whether or not the individual was in a given dataset, up to a statistical bound. Differential privacy is formally defined as follows \cite{10.1007/11681878_14}: given two neighboring datasets $D$ and $D'$, where neighboring datasets differ by either the absence or presence of exactly one row, a randomized algorithm $M$, and a subset $S$ of the possible outcomes of $M$, then:

\begin{equation}
\label{eq:differential_privacy}
\mathbb{P}(M(D) \in S) \leq e^{\epsilon}\mathbb{P}(M(D') \in S)
\end{equation}
where $\epsilon$ is the privacy bound, the lower the $\epsilon$ value, the greater the privacy. This strong privacy guarantee given by \cref{eq:differential_privacy} can be relaxed by introducing an additive $\delta$ parameter (typically very small), for which the $\epsilon$ bound does not hold. This relaxation is called $(\epsilon, \delta)$ differential privacy, and is obtained by updating \cref{eq:differential_privacy} as follows:

\begin{equation}
\mathbb{P}(M(D) \in S) \leq e^{\epsilon}\mathbb{P}(M(D') \in S) + \delta
\end{equation}

Building on this statistically rigorous definition of privacy, several works have introduced methods for training DP models. Abadi et al \cite{Abadi:2016:DLD:2976749.2978318} introduce the concept of a moments accountant, which provides tighter bounds on the composition of several DP operations, allowing for the creation of a DP stochastic gradient descent algorithm. As in other methods, the privacy guarantees are obtained by adding sampled noise to the algorithm $M$, in this case to the gradient calculated during the backpropagation step.

Zhang et al \cite{DBLP:journals/corr/abs-1801-01594} apply the same principle when training the discriminator network in GANs, thereby obtaining a differentially private GAN. We will utilize this adaption to GAN training later in our work.  Finally, both Esteban et al \cite{esteban2017realvalued} and Beaulieu-Jones et al \cite{doi:10.1161/CIRCOUTCOMES.118.005122} train versions of their models using DP SGD, and Esteban et al perform a similar data utility analysis as we will conduct on our model later in this work.\\[1ex]

To summarize, our contribution in this work is fourfold: (i) to the best of our knowledge, our model is the first that is able to generate heterogeneous (mixed data type) EHR records of high quality, which include administrative/demographic, as well as diagnosis and procedure code data; (ii) we match or outperform existing patient readmission classification methods in terms of AUROC, AUPRC and accuracy; (iii) our model is able to generate differentially private, heterogeneous datasets, and (iv) we show that our model does not introduce any bias into the generated data.

\section{Methods}
\label{sec:Methods}

\subsection{Data Description and Pre-processing}
\label{subsec:data}

Before detailing our generative model, we will first describe our dataset, as well as the data pre-processing steps that were carried out. For this work, we use the New Zealand National Minimal Dataset \cite{NZDataSet}. This data set consists of de-identified EHRs from the New Zealand health care system, which includes dense administrative features such as patient age, gender, and length of stay, as well as sparse features such as diagnosis codes. These features follow several different distributions: diagnosis and procedure codes are considered individual Bernoulli random variables, features such as admission month or department are categorical, age and length of stay are non-standard. This is therefore a heterogeneous, or mixed-type data set.

Several pre-processing steps were carried out before model training. In this work we consider only the years 2012 to 2017 (five years of training data, with one year used as the test set), so data from other years was removed. Furthermore, as our downstream task under consideration is returning patient prediction, we balanced the remaining dataset using this feature as a label and removed all rows in which the patient died. Finally, all continuous features were normalized to [0, 1] and categorical features were bucketized.

Next, to reduce the sparsity of the data, the ICD10 diagnosis codes were grouped into main diagnosis and additional diagnosis features, instead of having one feature per ICD10 code. One main diagnosis feature is created for each alphabetical diagnosis category, and one additional diagnosis column is created for each alphabetical and major number diagnosis category (e.g. one feature for ICD codes beginning with A1, B2). These diagnosis columns represent the presence of a diagnosis, each admission record has exactly one main diagnosis, but zero or more additional diagnoses. ICD10 procedure codes are grouped together after the 3rd digit. Finally, the real-valued features in the dataset are decomposed using Gaussian Mixture Models, as is done in the CTGAN \cite{xu2019modeling}. After preprocessing, the dataset contains a total of 2,873,466 rows and 828 columns. The years 2012-2016 are used for training (2,388,060 rows), and data from 2017 is used as the test set (485,406 rows).

\subsection{Model Description}

\begin{figure}
	\centering
	\includegraphics[width=\textwidth]{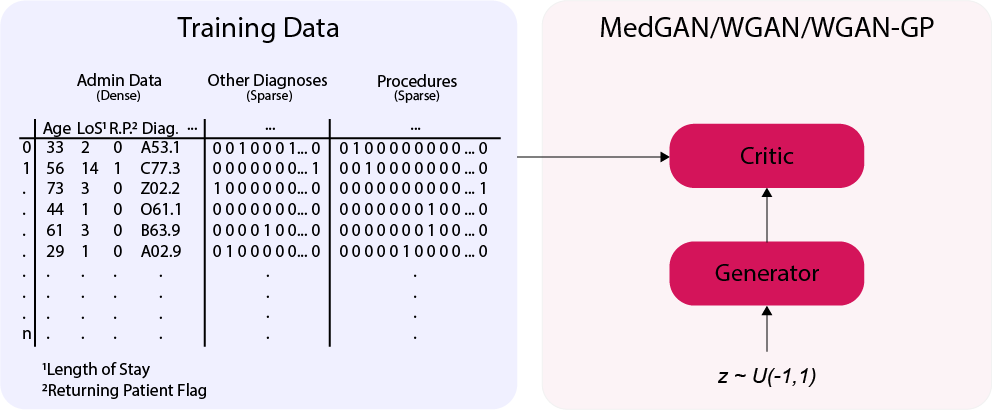}
	\caption{Our proposed model's architecture. The EHR training data is fed to the critic network of the MedGAN/WGAN/WGAN-GP model, along with the output of the generator network. The goal of the critic network is to learn a function maximizing the distance between the real and generated data distributions.}
	\label{fig:architecture}
\end{figure}

The main focus of this work is to generate synthetic EHR data using a GAN. In this section, we will describe how our model is configured and trained. Part of this process involved investigating the various GAN formulations described in \cref{subsec:GANs} and determining which one performed best for our task. We trained and evaluated three different models, the first is an implementation of MedGAN \cite{choi:2017} adapted to use our dataset, the second model is a standard WGAN model \cite{arjovsky:2017}, and the third model is a standard WGAN-GP model \cite{DBLP:journals/corr/GulrajaniAADC17}. \Cref{fig:architecture} gives a graphical representation of this system. We also performed some preliminary investigation into using the CTGAN model \cite{xu2019modeling} as well, however the large number of discrete features in our dataset led to markedly increased training time, as well as poorer results. Therefore, the CTGAN itself is not included as a potential model. However, as previously mentioned, we do utilize the GMM decomposition of real-valued features introduced by it in all three of our investigated models.

Given that the dataset described in \cref{subsec:data} is tabular in nature, we use fully connected neural networks to implement both the generator and discriminator in all of our models. In order to keep the size of the resulting network down (it can be trained on a normal desktop computer with/without a discrete GPU), we have limited both networks to 3 layers of hidden units. 

The generator network starts with a random input $z \sim U(-1, 1)$ of dimension 100, which is fed forward through three hidden layers of size [128, 256, 512] to the final output layer with a dimension of 828. All hidden layers use the ReLU activation function, the final layer uses the sigmoid function.

The discriminator network also uses three hidden layers, which are symmetric to the generator. Therefore, it begins with an input layer of dimension 828, feeding forward through hidden layers of size [512, 256, 128] before finally ending in one node which represents either the probability that the current sample was drawn from the training data (in the first model) or a critic score value used to approximate the Wasserstein-1 metric in the last two models. The hidden layers again use ReLU. The last layer in the MedGAN model uses sigmoid, in the WGAN models the last layer does not have an activation function.

\subsection{Model Selection}
\label{subsec:model_selection}

The process used to evaluate the three models and identify the best candidate involves comparing the fidelity of the generated data to the training data. The generated data fidelity is determined by comparing the features between the generated and training datasets by distribution. Features given by Bernoulli random variables are compared according to the maximum likelihood estimate of the $p$ parameter. The categorically distributed random variables are similarly compared using their estimated $p_{i}$ parameters, where $i \in K$, the set of possible categories for each random variable. Due to the bucketizing carried out in the pre-processing steps, it would theoretically be possible for the generated data to generate samples that belong to more than one category in any categorical feature (consider a sample generated where the binary columns for admission day of the week were 1 for Monday and Tuesday). We check that this has not occurred by summing these $p_{i}$, for each category this should equal one. Finally, in order to compare the overall data sets (including non-standard features such as age or length of stay), we compare the Frobenius norm of samples from both sets. The closer the generated data matches the training data, the closer these values will be.

Note that MedGAN includes many architectural features, some of which may have been necessary for the authors' datasets, but may not be required for our dataset. A full ablation study was performed on our MedGAN implementation in order to determine which combination of features performs best in our use case, after which a grid search over the various hyperparameters was performed. The best performing model was then chosen as our first model; in particular we ended up disabling batch normalization and generator shortcuts.

WGAN and WGAN-GPs are less dependent on hyperparameter selection for good performance, so we have used the recommended learning rate and optimizers given in the respective papers.

\begin{table}
	\caption{Data Fidelity results. Data Fidelity metrics are the average divergence from the training dataset for each random variable type in the input space, as well as the Frobenius norm divergence. All metrics are reported as the 95\% confidence interval based on three full experiment repetitions. Lower scores are better; the best model results are bolded. The DP model uses $\epsilon = 1$ and $\delta = 10^{-5}$.}
	\label{tbl:model_selection_results}
	\centering
	\resizebox{.99\columnwidth}{!}{
	\begin{tabular}{llll}
		\toprule
		& \multicolumn{3}{c}{Data Fidelity} \\
		\cmidrule(r){2-4}
		Model & Bernoulli & Categorical & Frobenius \\
		\midrule
		MedGAN Based Model & 0.0062 $\pm$ 0.000401 & 0.00557 $\pm$ 0.00128 & 8.41 $\pm$ 7.63 \\
		\midrule
		WGAN Model & \textbf{0.000755 $\pm$ 0.0000433} & 0.000954 $\pm$ 0.000203 & 7.7 $\pm$ 1.95 \\
		\midrule
		WGAN-GP Model & 0.000811 $\pm$ 0.00000726 & \textbf{0.000645 $\pm$ 0.0000362} & \textbf{2.71 $\pm$ 2.74} \\
		\bottomrule
		WGAN-GP w/ DP & 0.00539 $\pm$ 0.000548 & 0.00538 $\pm$ 0.00109 & 33.8 $\pm$ 33.9 \\
		\bottomrule
	\end{tabular}
	}
\end{table}

\Cref{tbl:model_selection_results} shows the data fidelity metrics of our three models after training. Both WGAN and WGAN-GP outperform the MedGAN based model, and out of those two the WGAN-GP model performs best on two of the three metrics. The average divergence from the baseline values Bernoulli features match to within an average divergence of 0.000811, and the categorically distributed features match to within an average divergence of 0.000645. Although noisier, the Frobenius norm divergence is also relatively low. As a further sanity check, for each categorical random variable, the probability of each category $p_{i}$ is summed, and the average difference from 1 is 0.00002143. Therefore, for the rest of this work, we will use the WGAN-GP generative model for both the DP application and the following experiments.

\subsection{Application of Differential Privacy}

\setlength{\intextsep}{0pt}
\begin{wrapfigure}[18]{r}{0.5\textwidth}
	\centering
	\includegraphics[width=0.5\textwidth]{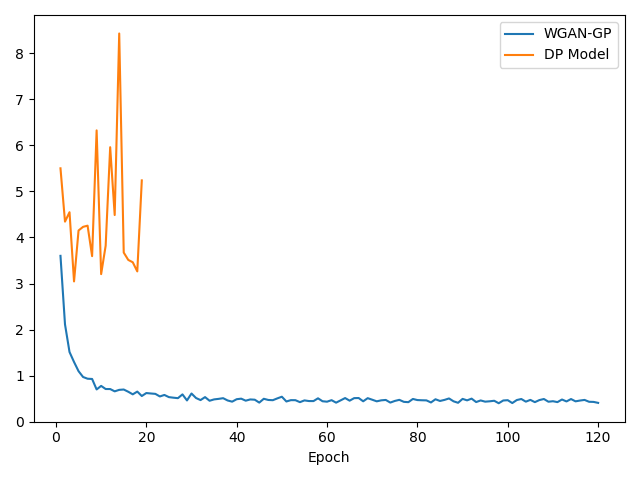}
	\caption{Generated data quality during training. The WGAN-GP (blue) progresses reliably to convergence, but the DP model (orange) is noisier. The DP model stops training after the 20th epoch due to having exhausted its privacy budget.}
	\setlength{\intextsep}{12pt}
	\label{fig:training_history}
\end{wrapfigure}

Building on the selected model, we apply the method described in \cref{subsec:differential_privacy} to create a DP version of our generative model. In particular, we retrain the model using a DP aware Adam optimizer, recently published by TensorFlow \cite{TFPrivacy}. The model is iteratively trained, and after processing each batch, the privacy accountant is updated as in \cite{Abadi:2016:DLD:2976749.2978318}. When the privacy budget has been exhausted, the training is stopped. Unfortunately, the noise added during the optimization process prevents the training process from smoothly converging. \Cref{fig:training_history} illustrates this problem.\\[1ex]

In order to ensure the best final result, the model parameters are saved after each training epoch and once the training process has exhausted the privacy budget, the best performing parameter set is used. The same data fidelity metrics are used to evaluate the quality of the DP EHR data, and are also listed in \cref{tbl:model_selection_results}. In this instance, we generated the data using $(1, 10^{-5})$ DP. The Bernoulli features have an average divergence of 0.00539 from the training data, and the categorical features match to an average divergence of 0.00538. For each categorical variable, the average divergence of the category probabilities from 1 is 0.00445229. The Frobenius norm divergence between the distribution samples is 33.8. These results, though less similar than the WGAN-GP model, still match the training distribution fairly well.

\setlength{\intextsep}{12pt}

\begin{figure}[t]
	\centering
	\includegraphics[width=1.0\textwidth]{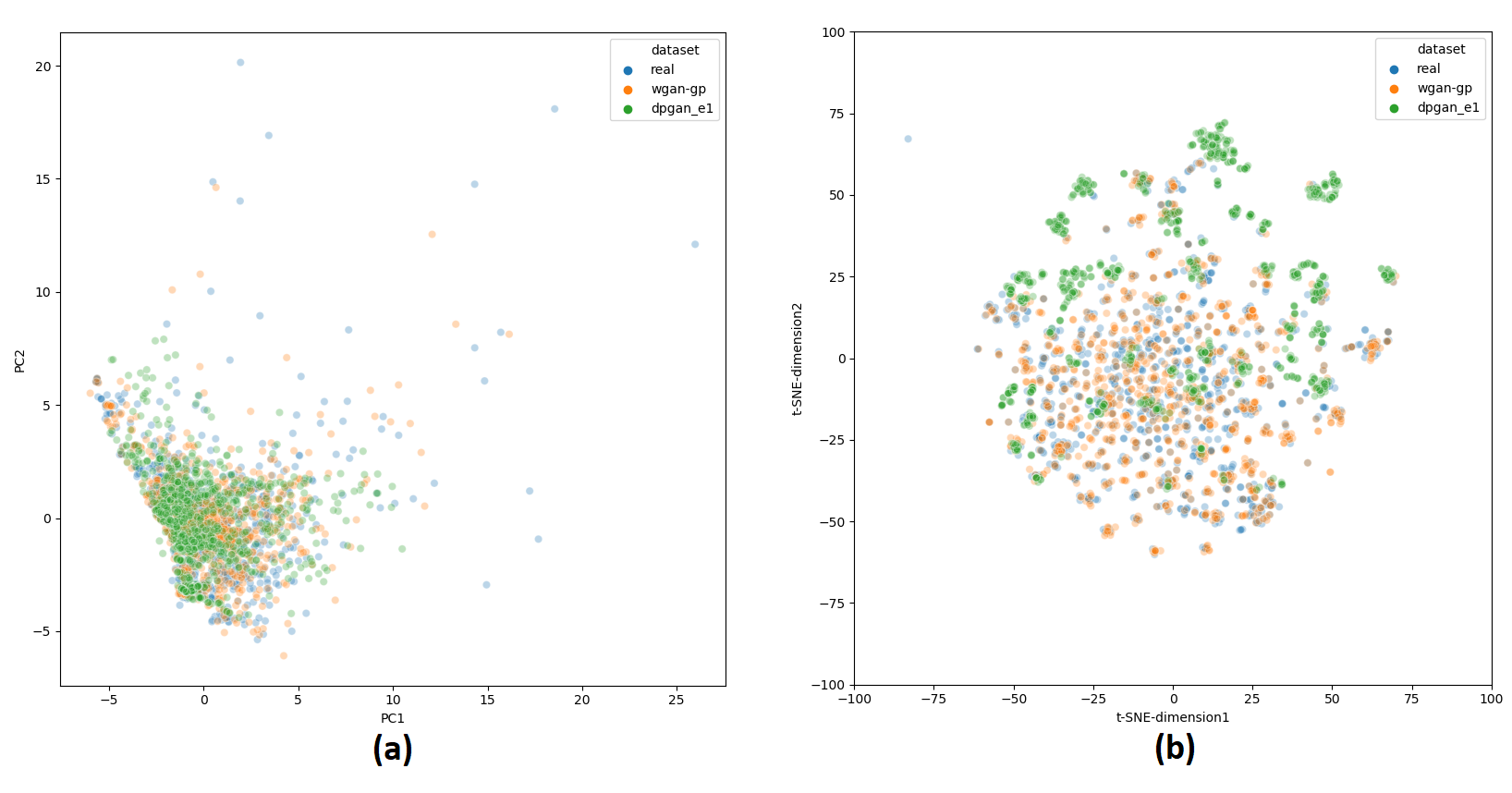}
	\caption{Plots of feature reduction comparing WGAN-GP and WGAN-GP w/ DP synthetic EHR data to baseline using (a) PCA and (b) t-SNE}
	\label{fig:pca_plots}
\end{figure}

\section{Experiments}
\label{sec:experiments}

The goal of this work is to generate EHR data which can be used for downstream tasks. To evaluate how well our generated data is suited for these tasks, we perform a set of experiments focused on using the generated EHR data to train binary classifiers. These classifiers are trained to predict the risk of a patient's early readmission to a hospital, as described in the introduction and \cref{subsec:readmission_classification}. The classifiers themselves are deep, fully connected neural networks consisting of [256, 128, 64] hidden layers of units using the ReLU activation function. Due to the fact that DP released datasets are immune to post-processing privacy penalties (as long as the original dataset is in no way accessed) \cite{Dwork:2014:AFD:2693052.2693053}, these binary classifiers can be trained to convergence without worry about incurring any additional privacy loss. The performance of the classifiers is evaluated using the AUROC, AUPRC, and accuracy metrics, and are referred to as the data utility metrics of the generated EHR data. If the generative models have learned the training distribution well, the performance of the classifiers should be similar to that of classifiers trained using the training dataset. Evaluation is performed on both the WGAN-GP and DP models. The results of these experiments are given below, and analyzed in \cref{subsec:experiment_analysis}. Further analysis of the impact of enabling DP follows in \cref{subsec:dp_impact}.

\section{Results}

The results obtained from the binary classifiers are presented in \cref{tbl:experiment_results}. The complete experiment was repeated three times and the results are reported as the 95\% confidence interval. Metrics considered are the area under the Receiver Operating Characteristic (AUROC) and precision-recall (AUPRC), as well as the classification accuracy. Finally, it is worth taking a qualitative look at the data that is produced by our models. To do this, we analyze the synthetic data from both the WGAN-GP and DP model using PCA and t-SNE and plot the first two principal components (PCA) or dimensions (t-SNE) for 1000 samples from each dataset in \cref{fig:pca_plots}. We would expect the point clouds to overlap relatively well if the generative models have learned the training data distribution well.

\begin{table}[b]
	\caption{Experiment Results. Data Utility metrics are the AUROC and AUPRC values, as well as the classification accuracy reported as the 95\% confidence interval based on three full experiment repetitions. The DP configuration uses $\epsilon = 1$ and $\delta = 10^{-5}$. Higher scores are better.}
	\label{tbl:experiment_results}
	\centering
	\resizebox{.8\linewidth}{!}{
	\begin{tabular}{llll}
		\toprule
        & \multicolumn{3}{c}{Data Utility} \\
		\cmidrule(r){2-4}
		Model & AUROC \,& AUPRC \,& Acc. \\
		\midrule
		Baseline & 0.8003 \, & 0.8245 \,& 0.7171 \\
		\midrule
		WGAN-GP & 0.761 $\pm$ 0.0028 \, & 0.782 $\pm$ 0.00206 \,& 0.694 $\pm$ 0.00335 \\
		\bottomrule
		WGAN-GP w/ DP & 0.661 $\pm$ 0.0147 & 0.679 $\pm$ 0.0269 & 0.603 $\pm$ 0.019 \\
		\bottomrule
	\end{tabular}
	}
\end{table}

\section{Analysis}

\subsection{Experiment Analysis}
\label{subsec:experiment_analysis}

The results in \cref{tbl:experiment_results}, as well as the data fidelity metrics shown in \cref{tbl:model_selection_results} demonstrate that both generative models are capable of producing heterogeneous EHR data which can be used to perform downstream binary classification, and most likely other downstream tasks as well. Compared to the training data baseline, which itself obtains an AUROC value comparable to or better than the models in \cref{subsec:readmission_classification}, when using the synthetic training data to classify real patient data from the test set, the AUROC, AUPRC and accuracy metrics are only marginally worse, 0.0393 (4.9\%), 0.0425 (5.1\%), and 0.0231 (3.2\%), respectively. This demonstrates that for this studied classification task, the synthetic data could viably be used in place of the real training data, given that it performs as good or better than previous works outlined in \cref{subsec:readmission_classification}.

In line with what has already been observed in terms of data fidelity, the DP synthetic EHR data has less data utility than WGAN-GP. Particularly, the AUROC is 0.1393 (17.4\%) lower than the baseline, the AUPRC is 0.1455 (17.6\%) lower, and similarly the accuracy is 0.1141 (15.9\%) lower. Despite the lower performance, these results are encouraging because they demonstrate that DP synthetic EHR data can still be used for downstream binary classification tasks. The obtained AUROC value of 0.661 is still significantly higher than the random guess baseline of 0.5 which demonstrates that when privacy guarantees are desired or required, using a model such as ours is a viable option.

Finally, looking at the qualitative comparisons in \cref{fig:pca_plots}, we observe that both the WGAN-GP and DP models are able to generate the "average" data well, but have less success generating outliers according to the PCA reduction. The t-SNE reduction shows that the WGAN-GP model's data matches the average case somewhat better than the DP model, this observation perhaps more closely matches the quantitative results that we obtained. In future research we would like to dig deeper into the outlier cases and try to determine a reason for this. For the DP model at least, it seems reasonable to assume that this is partially due to the noise being added during the training process. The already rare outliers' signal is further reduced by the addition of noise, and if too much noise is added the influence of these data points could be removed completely.

\subsection{Impact of Differential Privacy}
\label{subsec:dp_impact}

We have already observed one impact of enabling DP, namely that the training process becomes quite noisy and does not converge due to the noise added during optimization. Looking once again at \cref{fig:training_history}, it is obvious that the DP model not only fails to converge, but also doesn't get to train for as long as the WGAN-GP model, due to the privacy budget being exhausted. As a result we see that the obtained AUROC, AUPRC and accuracy values are 13.1\%, 13.2\% and 13.1\% worse compared to the WGAN-GP model.

\begin{figure}
    \centering
    \includegraphics[width=0.99\linewidth]{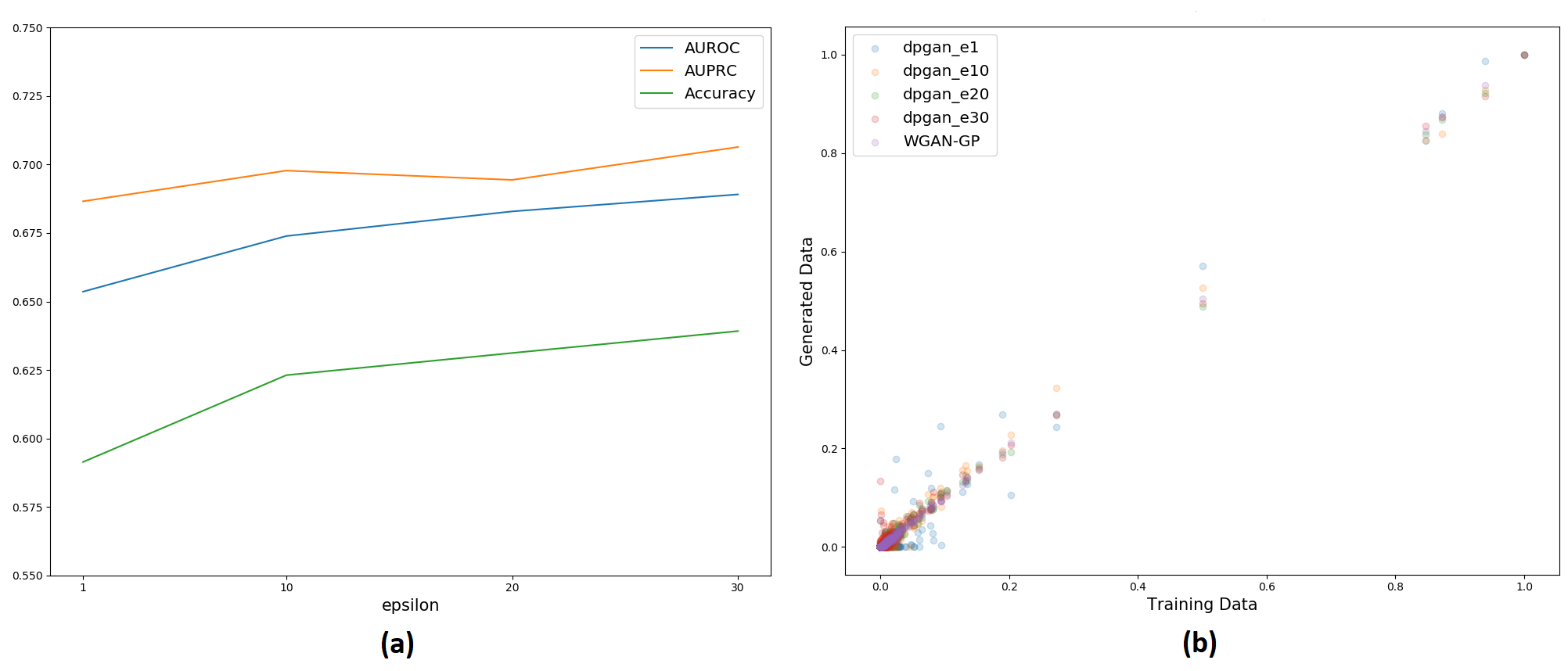}
    \caption{(a) Data utility plotted vs epsilon. Utility increases as privacy guarantees are lowered, (b) Bernoulli features for generated and training data. The closer the generated data is to the training data, the closer its points are to the diagonal}
    \label{fig:diff_priv}
\end{figure}

One implication of this could be that if the privacy budget was increased, we should expect to see improvements in the fidelity and utility of the generated EHR data. To test this hypothesis, we trained several additional DP models, with varying $\epsilon$ values: $\epsilon = 1, 10, 20, 30$. We then generated synthetic EHR data using these models and train binary classifiers as before. The data utility metrics produced by these classifiers are plotted in \cref{fig:diff_priv} (a). As we might expect, the general trend is that as the privacy restriction is loosened, the data utility increases. We expect that if this process were continued until $\epsilon = \infty$, the utility of that generated data would approximately equal the non-private data.

Another way to look at the impact of DP is to dig a little deeper into the data fidelity. \Cref{fig:diff_priv} (b) shows the $p$ value of the Bernoulli features of the various generated EHR datasets compared to the training data. As shown in the plot, the whole process is somewhat noisy, however the trend appears to be that as the privacy restriction is increased, the difference between generated and baseline $p$ values also increases.

\subsection{Sub-population Analysis}
\label{subsec:sub-population_analysis}

Fairness and bias in machine learning has emerged as a very important topic, particularly as our lives are impacted more and more by it \cite{barocas-hardt-narayanan}. A recent study by the National Institute of Standards and Technology \cite{NIST} revealed the extent to which facial recognition algorithms' performance can suffer when evaluated with particular sub-population groups, due to biases inherent in the training data. As one potential use case of our model is to create EHR datasets from a hospital's own dataset, we would like to understand whether or not biases may also be present in our data, and more importantly, whether or not any bias is introduced by our generative model.

To this end, we evaluate our classifiers trained using the synthetic EHR data from the WGAN-GP and DP models using sub-populations of the testing data set. Specifically, we test the male and female sub-populations, as well as by age groupings (0-18 years, 19-50 years and 51+ years). If our dataset contained more background information such as patient ethnicity it would be possible to do a more thorough analysis.

The results of this analysis are presented in \cref{tbl:sub-population_analysis}. The results of our baseline classifier should reveal if bias exists in our training dataset. What we observe is that there is almost no difference in performance between the male and female sub-populations, as the AUROC, AUPRC and accuracy metrics differ only by 0.5\%, 0.2\% and 0.7\% respectively. When looking at the age groups however, we do notice a trend where the 0-18 age group performs best in two of the three metrics (AUROC and accuracy), followed by the 19-50 age group and then the 51+ age group, which has the best AUPRC result. 

\begin{wraptable}{l}{0.6\linewidth}
	\caption{Sub-population analysis results for male/female and age groups 0-18, 19-50, 51+. Higher scores are better.}
	\label{tbl:sub-population_analysis}
	\centering
	\resizebox{.99\linewidth}{!}{
	\begin{tabular}{llll}
		\toprule
        & AUROC & AUPRC & Acc. \\
		\cmidrule(r){2-4}
		Model & \multicolumn{3}{c}{Female Sub-population} \\
		\midrule
		Baseline & 0.8029 & 0.8251 & 0.7206 \\
		\midrule
		WGAN-GP & 0.758 $\pm$ 0.00293 & 0.775 $\pm$ 0.00485 & 0.692 $\pm$ 0.00274 \\
		\midrule
		WGAN-GP w/ DP & 0.656 $\pm$ 0.0139 & 0.65 $\pm$ 0.029 & 0.598 $\pm$ 0.0156 \\
		\bottomrule
		\\
		& \multicolumn{3}{c}{Male Sub-population} \\
		\midrule
		Baseline & 0.7987 & 0.8232 & 0.7156 \\
		\midrule
		WGAN-GP & 0.749 $\pm$ 0.00111 & 0.773 $\pm$ 0.00112 & 0.682 $\pm$ 0.00242 \\
		\midrule
	    WGAN-GP w/ DP & 0.664 $\pm$ 0.00999 & 0.67 $\pm$ 0.00438 & 0.602 $\pm$ 0.0184 \\
		\bottomrule
		\\
		& \multicolumn{3}{c}{Age 0-18} \\
		\midrule
		Baseline & 0.8088 & 0.7945 & 0.7589 \\
		\midrule
		WGAN-GP & 0.752 $\pm$ 0.00438 & 0.739 $\pm$ 0.00362 & 0.725 $\pm$ 0.00136 \\
		\midrule
		WGAN-GP w/ DP & 0.682 $\pm$ 0.0145 & 0.622 $\pm$ 0.0393 & 0.607 $\pm$ 0.0252 \\
		\bottomrule
		\\
		& \multicolumn{3}{c}{Age 19-50} \\
		\midrule
		Baseline & 0.8021 & 0.8157 & 0.7241 \\
		\midrule
		WGAN-GP & 0.75 $\pm$ 0.0054 & 0.756 $\pm$ 0.00756 & 0.692 $\pm$ 0.00345 \\
		\midrule
		WGAN-GP w/ DP & 0.638 $\pm$ 0.0158 & 0.613 $\pm$ 0.037 & 0.588 $\pm$ 0.014 \\
		\bottomrule
		\\
		& \multicolumn{3}{c}{Age 51+} \\
		\midrule
		Baseline & 0.7889 & 0.8336 & 0.7001 \\
		\midrule
		WGAN-GP & 0.747 $\pm$ 0.00189 & 0.79 $\pm$ 0.00088 & 0.671 $\pm$ 0.00232 \\
		\midrule
		WGAN-GP w/ DP & 0.666 $\pm$ 0.0291 & 0.701 $\pm$ 0.0209 & 0.605 $\pm$ 0.0317 \\
		\bottomrule
	\end{tabular}
	}
\end{wraptable}

This mixed performance could indicate that some bias exists, particularly since the accuracy metric is almost 8\% worse in the 51+ age group when compared to the 0-18 age group. This is particularly interesting since there are more samples in the training data in the 51+ age group. However, it could also simply indicate that predicting hospital re-admission in elderly patients is a more difficult problem than for younger patients -- to fully understand this issue we would need to investigate the cases that were incorrectly classified and try to understand if there are any identifiable patterns.

Looking now at the classifiers trained using EHR data from the WGAN-GP model, we see the same patterns as in the baseline. That is, there is again a very small difference between the male and female sub-populations, ~1\% in all metrics. In the age groups, we again observe the 0-18 group with the best AUROC and accuracy performance, while the 51+ age group has the best AUPRC performance.

The classifiers using DP EHR data are a little noisier, which was expected. However, the male and female sub-populations differ by only a maximum of 3\% in the AUPRC metric. The trends that were observed before in the age groups mostly also hold, except that we observe a tighter range of performance in the accuracy metric than previously. Potential reasons for this include that the noise added during training is impacting all samples and thus limiting the impact of any one, as is the intent of DP. However, this may also have the side-effect that bias caused by the imbalance of samples belonging to sub-populations within a dataset is limited, since each sample has a diminished impact on the optimization process when compared to training without DP.

In general, this indicates that our models do not introduce any bias that was not present in the original dataset, and using DP could potentially mitigate inherent bias by virtue of reducing the impact of outlier data. However, we could not exhaustively test this given the lack of more specific features in our dataset.

\section{Conclusion}

To conclude, in this paper we have introduced a generative model capable of producing high quality synthetic EHR records using a large, high-dimensional and highly sparse heterogeneous dataset. Our model employs a Wasserstein GAN using the gradient penalty method, and this method was chosen after a rigorous, data-driven evaluation of three different models. We then extended this model by retraining it with a DP SGD optimizer, thereby creating a model capable of generating differentially private, heterogeneous EHR datasets. 

These two models were then used for a downstream binary classification task, and we found that the WGAN-GP model produces high quality data, very close to the baseline. In particular, we successfully used the synthetic data to train a classifier for which the lowest performing metric is within 5\% of the baseline. This data could also be used to supplement datasets where there is class imbalance, and in other instances where privacy is not a concern. Introducing DP incurs a larger performance penalty (17\% lower than baseline in the worst case), but we showed that it is still useful for training downstream classifiers. This is particularly valuable when data privacy is a key issue. Finally, we analyzed the performance of our models across the male and female sub-populations, and across age groups. We found no evidence of bias being introduced by our models, however there is some evidence that our training data could be biased towards producing better classification results for patients in the 0-18 age group.

We believe that differentially private EHR dataset generation could be a valuable tool to machine learning and health care, since it could allow the release of many more interesting datasets from hospitals around the world, broadening the amount of potential research avenues and perhaps in turn leading to many exciting discoveries.

\bibliographystyle{splncs04}
\bibliography{document}

\end{document}